\documentclass{article}
\usepackage{amsmath,graphicx,mlspconf}
\usepackage{multirow}
\usepackage{subcaption}
\usepackage{lipsum}
\usepackage{hyperref}
\toappear{2024 IEEE International Workshop on Machine Learning for Signal Processing, Sept.\ 22--25, 2024, London, UK}

\title{Normalizing energy consumption for hardware-independent evaluation}

\name{Constance Douwes, Romain Serizel}
\address{Université de Lorraine, CNRS, Inria, Loria, Nancy, France}

\begin{document}

\maketitle

\begin{abstract}
% In recent years, machine learning models have made impressive performance gains in signal processing. However, the environmental impact of those models, particularly during resource-intensive training, has become a growing concern. Therefore, quantifying the energy consumption associated with ML model training and comparing it with performance is crucial for more sustainable machine learning evaluation practices. One of the key challenges is to find an energy metric that remains consistent across various hardware to ensure fair comparisons between systems. In this paper we explore various normalization strategies and apply our methodology on multiple machine learning architectures for sound classification task. We evaluate the energy consumption for training those network on different hardware and we show that by normalizing with two of the system energies we can provide fair energy comparisons as well as inclusing computational metrics. 

The increasing use of machine learning (ML) models in signal processing has raised concerns about their environmental impact, particularly during resource-intensive training phases. In this study, we present a novel methodology for normalizing energy consumption across different hardware platforms to facilitate fair and consistent comparisons. We evaluate different normalization strategies by measuring the energy used to train different ML architectures on different GPUs, focusing on audio tagging tasks. Our approach shows that the number of reference points, the type of regression and the inclusion of computational metrics significantly influences the normalization process. We find that the appropriate selection of two reference points provides robust normalization, while incorporating the number of floating-point operations and parameters improves the accuracy of energy consumption predictions. By supporting more accurate energy consumption evaluation, our methodology promotes the development of environmentally sustainable ML practices.
\end{abstract}
\begin{keywords}
Machine learning, energy consumption, normalization, GPU, FLOPs, signal processing.
\end{keywords}
\section{Introduction}

In the field of signal processing, deep learning (DL) has seen a widespread adoption with applications in various domains, such as music generation \cite{agostinelli2023musiclm}, speech recognition \cite{radford2023robust} and sound event detection (SED) \cite{li2023ast}. However, its growing popularity has drawn attention to a major problem : the significant amount of compute associated with learning and running deep learning models \cite{amodei2018ai, sevilla2022compute,thompson2007computational}. Notably, in the field of natural language processing (NLP), Strubell et al. \cite{strubell2019energy} have highlighted the colossal environmental and financial impact of training such large language models. This research contributes to an emerging field known as Green AI \cite{schwartz2020green}, which seeks to measure and mitigate the energy consumption of AI technologies, with new tools for estimating both energy consumption and carbon emissions \cite{schmidt2021codecarbon, anthony2020carbontracker, lacoste2019quantifying}. 

Even though models used in audio processing are smaller than those used in NLP, they still present similar problems~\cite{douwes2021energy, parcollet2021energy}, and efforts have been made balance energy efficiency with performance. For example, Douwes et al.~\cite{douwes2023quality} performed a detailed analysis of generative models for audio synthesis, exploring the trade-offs between energy consumption and audio quality. In speech recognition, Parcollet and Ravanelli \cite{parcollet2021energy} showed that small performance improvements often have extremely high energy costs. Similarly, Serizel et al.~\cite{serizel2023performance} investigated the relationships between energy consumption, training time, GPU type and performance when training SED systems. In addition, Ronchini et al. \cite{ronchini2023performance} balanced performance and energy based on an energy-weighted metric \cite{ronchini2022description} to evaluate SED systems across different hardware types. This issue is particularly critical in the context of cross-hardware comparison, where a consistent and comparable measurement of energy consumption is required. While it is common practice in research to test a system's performance and compare it with others, comparing energy consumption presents significant challenges due to the heavy dependence on the hardware used. Even if systems perform similarly, they may use different amounts of energy depending on the hardware they are trained on. Therefore, it is imperative to develop methods for normalizing energy metrics, ensuring independence from hardware and isolating the influence of the model's energy consumption. For example, from 2022, the DCASE Challenge Task 4 \cite{ronchini2022description} requests participants to report their energy consumption along with a reference consumption in order to normalize energy consumption. However, no experimental analysis of this normalization strategy has been conducted to prove its effectiveness.

A potential alternative approach to assessing energy efficiency is to use computational metrics, such as the number of floating-point operations (FLOPs), as proxies. This method has been explored by Asperti et al. \cite{asperti2021dissecting}, where they estimated the energy consumed by convolutional networks during inference on GPU using a slightly modified computation of FLOPs. However, their study was limited to CNN architectures, which raises questions about the applicability of this metric to other neural network architectures. In particular, Ronchini et al.  \cite{ronchini2023performance} showed that the relationship between FLOPs and energy consumption is not straightforward without prior knowledge of the network. Another approach to energy estimation is to predict per-layer energy consumption given layer characteristics, as done by Getzner et al. \cite{getzner2023accuracy}. The total consumption of the network is then the sum of all the individual predicted energy contributions of the layers. Again, detailed knowledge of the architecture is required to accurately estimate the total consumption. Although these methods give promising results, they are heavily dependent on specific equipment and do not address training. Yet, as ML audio researchers, the majority of our energy consumption lies in the training phase and should not be neglected.

In this paper, we focus on developing a methodology to compare the energy consumption of the training processes independently of the hardware used, with minimal information on the neural network architecture. As an initial study, we measure the energy consumed during the training of four neural network architectures (MLP, CNN, RNN, CRNN) designed for audio tagging tasks on four different GPUs. We first study the influence of the number of reference point to normalize the energy consumption. We show that by strategically choosing those points, a strong linear regression fit can be achieved. We then explore different regression models to determine which one best fits the energy consumption on two hardware. We consider linear, polynomial and support vector regression (SVR). Our results show that the most appropriate regression model varies depending on the specific hardware pair. Finally, we integrate computational metrics such as the number of FLOPs and parameters to the linear regression. This hybrid approach provides robust predictions that adapts to diverse hardware configurations.

\section{Experimental setup}

Our goal is to normalize the energy consumption for training neural network models on different hardware. As a proof of concept, we work on an audio tagging task, and record the energy for training different types and sizes of architectures.\footnote{\url{https://github.com/ConstanceDws/toolbox_energy}}

\paragraph*{Task description.} Audio tagging consists in assigning tags (one or many) to an audio signal without any additional temporal information. In this experiment, we use the DESED dataset \cite{turpault2019sound}, a well-known resource in the sound event detection community. This dataset consists of 10-second audio clips containing sound from domestic environments. We focus on the subset of real recordings, which provide a reliable 10-class annotation. We convert these recordings into mel-spectrogram representations using 128 bands, with an FFT size of 2048 and a hop size of 256. We use the first 64 frames as input, which approximately corresponds to taking the first 1 second of the audio signal. This drastically changes the performance of the model but also reduces the complexity of the system and allows for lighter experiments, as we do not focus on performance but only on energy.

\paragraph*{Models.} We implement four neural network architectures: multi-layer perceptron (MLP), convolutional neural network (CNN), recurrent neural network (RNN), and convolutional recurrent neural network (CRNN). For the MLP, we implement a series of linear layers followed by ReLU activation functions. For the CNN, we adopt a sequence of Conv2d, ReLU and MaxPool2d layers. For the RNN, we use GRU cells and for the CRNN, we combine both CNN and RNN. All implementations are completed with a final linear layer and a sigmoid activation function that outputs a probability vector for the 10 classes. For each architecture, we systematically increase the number of layers and adjust the hidden sizes per layer, gradually scaling up to reach the full GPU memory capacity and utilization, resulting in 43 models. We present the summary of all the configurations tested in Table \ref{tab:model}. We intentionally chose those configurations to achieve meaningful variations in the number of FLOPs without conducting redundant experiments.

\begin{table}[t]
    \centering
    \begin{tabular}{ccc}
        %\hline
        \textbf{Model} & \textbf{Num Layers} & \textbf{Hidden Sizes} \\
        \hline
        \multirow{3}{*}{MLP} & 1 & 512, 1024, 2048 \\
        & 4 & 1024, 2048, 4096 \\
        & 6, 10, 16, 32 & 4096 \\
        \hline
        \multirow{3}{*}{CNN} & 1 & 128, 256, 512, 1024 \\
        & 2 & 128, 256, 384, 512, 768, 1024 \\
        & 6 & 384, 768 \\
        \hline
        \multirow{3}{*}{RNN} & 1 & 128, 512, 1024, 2048 \\
        & 4, 6& 1024, 2048 \\
        & 2, 10, 14 & 2048 \\
       \hline
        \multirow{3}{*}{CRNN} & [1,1], [2,1], [1,2] & [64,64], [256,64], [512, 256]  \\
        & [2,2] & [728, 256]\\
        & [1,2], [2,2] & [1024, 256] \\
    \end{tabular}
    \caption{Summary of all the configurations tested in our experiment. For each number of layer, we tested different hidden sizes. For CRNN, the configurations first indicate the convolutional layers and then the recurrent layers.}
    \label{tab:model}
\end{table}

\paragraph*{Training.} Our experiments differ from the traditional search for accuracy measures. Instead, we focus our analysis on the energy consumption associated with training neural networks. To do this, we trained all models for 10 epochs on four different GPU types. The specifications for each GPU are summarized in Table \ref{tab:hardware} and are referred to throughout the paper by their respective names: RTX, GTX, T4 and A40. We also include the Thermal Design Power (TDP) of each GPU, which indicates the maximum heat generated at peak load, providing an insight into the power requirements of each unit. We chose a common batch size of 8, use the cross-entropy function as the criterion, and set the learning rate at $10^{-3}$ with an ADAM optimiser \cite{kingma2014adam}. We deliberately omit any validation steps in the training routine in order to attribute the energy measurements only to the training process. 

\begin{table}[htbp]
\centering
\begin{tabular}{lr}
\textbf{Hardware} & \textbf{TDP} \\ \hline
NVIDIA GeForce RTX 2080 Ti (11 GiB) & 250 W \\ 
NVIDIA GeForce GTX 1080 Ti (11 GiB)  & 250 W\\ 
NVIDIA Tesla T4 (15GiB) & 70 W \\ 
NVIDIA A40 (45 GiB) & 300 W \\ 
\end{tabular}
\caption{Specifications of the GPUs used in the study, including each unit's Thermal Design Power (TDP).}
\label{tab:hardware}
\end{table}

\paragraph*{Energy and computational costs.}
We monitor the energy consumption for training using CodeCarbon \cite{schmidt2021codecarbon}, which provides detailed energy consumption for the three components of the computing system: GPU, CPU and RAM. We focus exclusively on the GPU consumption, as this is the primary energy drain in our experiments. We also compute the number of FLOPs using the deepspeed profiler ~\cite{rasley2020deepspeed} to quantify the number of forward pass operations accurately. We also report the number of parameters of each configuration. 

\section{Results}

We start with a study of the normalization strategy proposed by Ronchini et al. \cite{ronchini2023performance} and extend to a quantitative analysis of the influence of the number of reference points chosen for the regression used as normalization, followed by a study of the influence of the type of regression, and the inclusion of computational metrics to the regression.

\subsection{Influence of the number of reference points}

% \begin{figure*}[t]
%     \centering
%     \begin{subfigure}[b]{\textwidth}
%         \includegraphics[width=\linewidth]{plot/grue_gruss_exp1.pdf}
%         \caption{T4-A40}
%     \label{fig:T4_A40_exp1}
%     \end{subfigure}
%     \begin{subfigure}[b]{\textwidth}
%         \includegraphics[width=\linewidth]{plot/grue_graffiti_exp1.pdf}
%         \caption{T4-RTX}
%         \label{fig:T4_RTX_exp1}
%     \end{subfigure}
%     \caption{Normalization of energy consumption using different reference points across two GPU pairs, T4-A40 and T4-RTX. From left to right: unnormalized energy consumption (in kWh), normalization using a low-energy reference model, normalization using a high-energy reference model, and normalization using both low and high-energy reference models. The x=y line indicates optimal normalization.}
%     \label{fig:combined_figures_exp1}
% \end{figure*}

\begin{figure}[t]
    \centering
    \includegraphics[width=\linewidth]{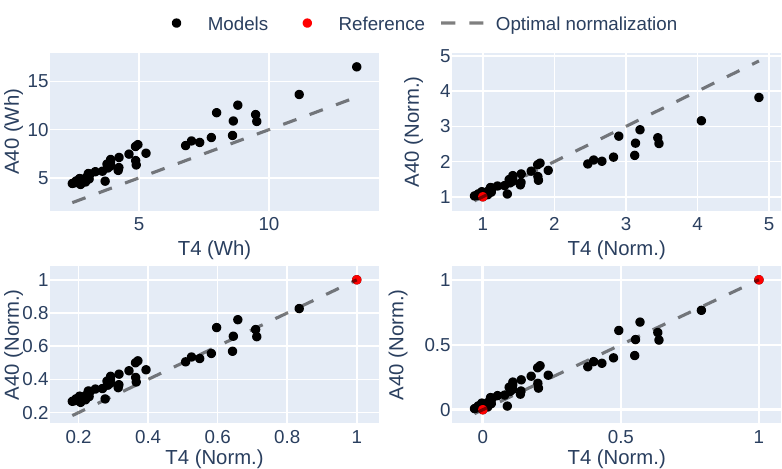}
    \caption{Normalization of energy consumption using different reference points for the T4-A40 GPU pair.}
    \label{fig:T4_A40_exp1}
\end{figure}

\begin{figure}[t]
    \centering
    \includegraphics[width=\linewidth]{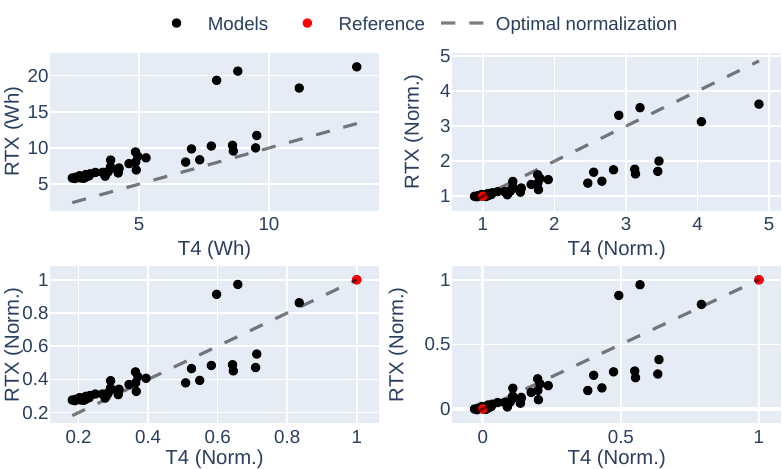}
    \caption{Normalization of energy consumption using different reference points for the T4-RTX GPU pair.}
    \label{fig:T4_RTX_exp1}
\end{figure}

% Normalization of energy consumption using different reference points across two GPU pairs, T4-A40 and T4-RTX. From left to right: unnormalized energy consumption (in kWh), normalization using a low-energy reference model, normalization using a high-energy reference model, and normalization using both low and high-energy reference models. The x=y line indicates optimal normalization.

\begin{figure}[t]
    \centering
    \begin{subfigure}[b]{0.49\linewidth}
        \includegraphics[width=\linewidth]{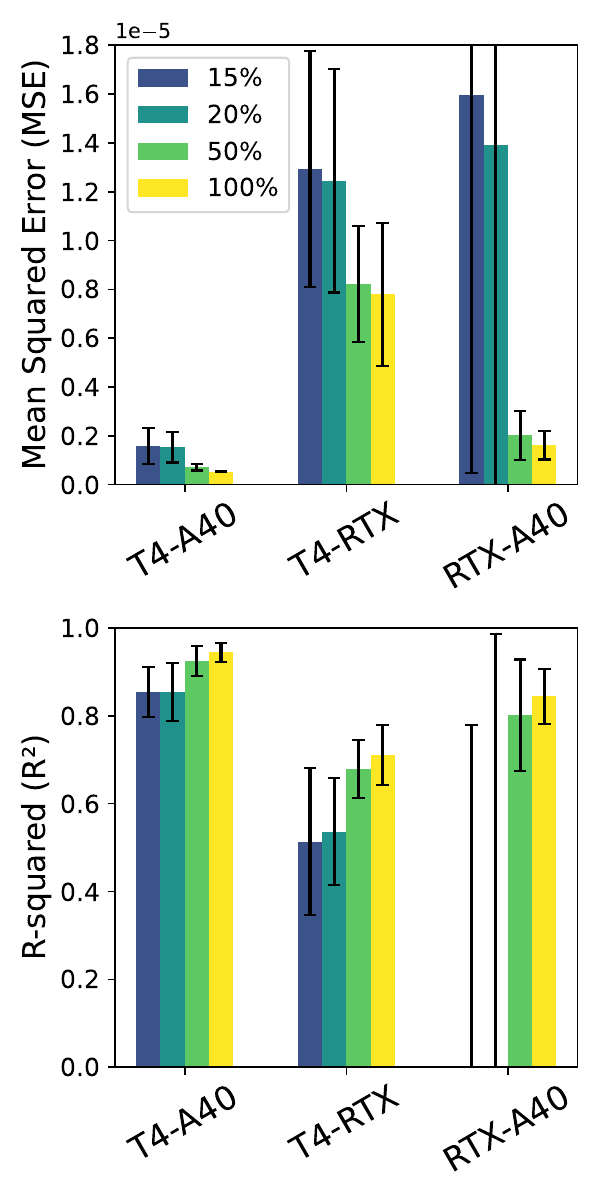}
        \caption{Random sampling}
        \label{fig:random}
    \end{subfigure}
    \begin{subfigure}[b]{0.49\linewidth}
        \includegraphics[width=\linewidth]{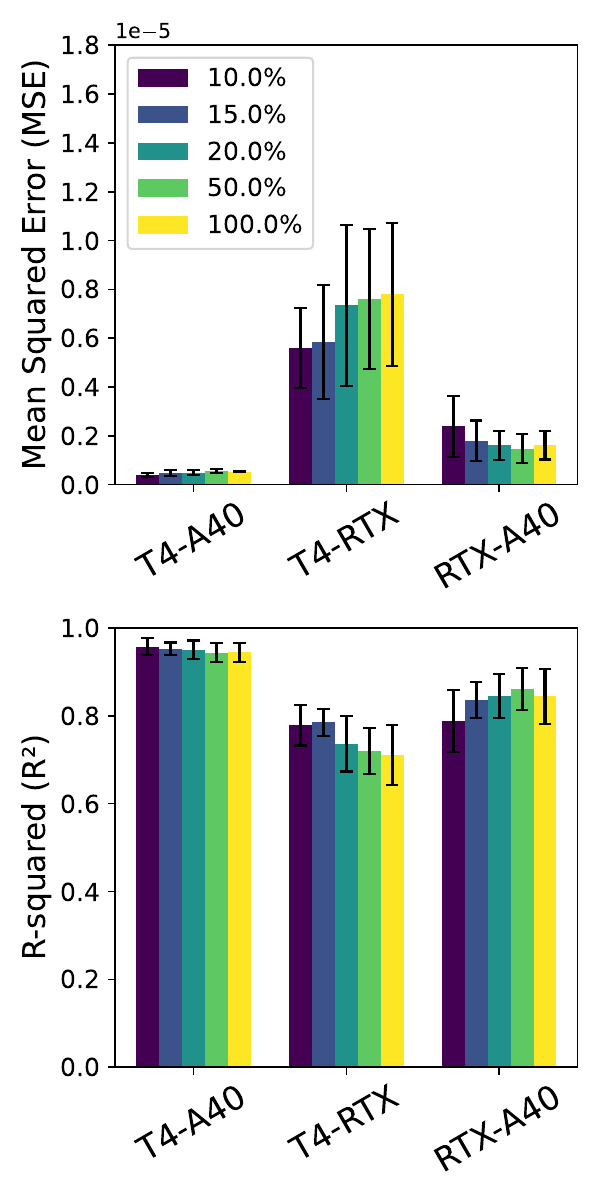}
        \caption{Min-max sampling}
        \label{fig:extrem}
    \end{subfigure}
    \caption{Impact of reference point selection on the linear regression between GPU pairs using different data sampling strategies.}
    \label{fig:sampling}
\end{figure}

We explore here the normalization strategy from Ronchini et al. \cite{ronchini2023performance}, which involves selecting a reference model and recording its energy consumption on each GPU. This approach assumes that the ratio of energy consumption between any given model and the reference model should remain consistent across different GPU types. Figures \ref{fig:T4_A40_exp1} and  \ref{fig:T4_RTX_exp1} show the results of this normalization experiment for T4-A40 and T4-RTX GPU pairs respectively. The top left plots show the energy consumption without any normalization. On the top right the graphs show the normalization using a low energy model as the reference, and on the bottom left a high energy model as the reference. The last graphs at the bottom right present our dual reference approach, where both a low and a high energy model are used as reference points. The closer the results are to the optimum line (in dashed), the more accurate the normalization is. We can see that the normalization is significantly affected by the choice of the reference model. Using a low-energy model as the reference effectively normalizes the energy consumption of other low-energy models but reduces normalization accuracy for high-energy models, and vice versa. Our strategy, which uses dual reference points, achieves a closer approximation to the ideal linear relationship between T4-A40. However, for the T4-RTX pair, while the dual-reference approach corrects the inclination of the normalization for low and high models, the linear regression still appears non-optimal. Note that this dual approach is a two-point regression.

% This principle is mathematically expressed as:

% \begin{equation}
% \left[ \frac{\text{kWh}_{\text{model}}}{\text{kWh}_{\text{reference}}} \right]_{\text{GPU X}} = \left[ \frac{\text{kWh}_{\text{model}}}{{\text{kWh}_{\text{reference}}}}\right]_{\text{GPU Y}}
% \label{eq:kwh}
% \end{equation}

% where $\text{kWh}_{\text{model}}$ and $\text{kWh}_{\text{reference}}$ represent the kilowatt-hours consumed by the model and the reference model, respectively. $\text{GPU X}$ and $\text{GPU Y}$ denote different types of GPU hardware, illustrating that energy efficiency comparisons are intended to be hardware-independent. 

% [], verifying equation (\ref{eq:kwh}).

\begin{figure}[t]
    \centering
    \includegraphics[width=\linewidth]{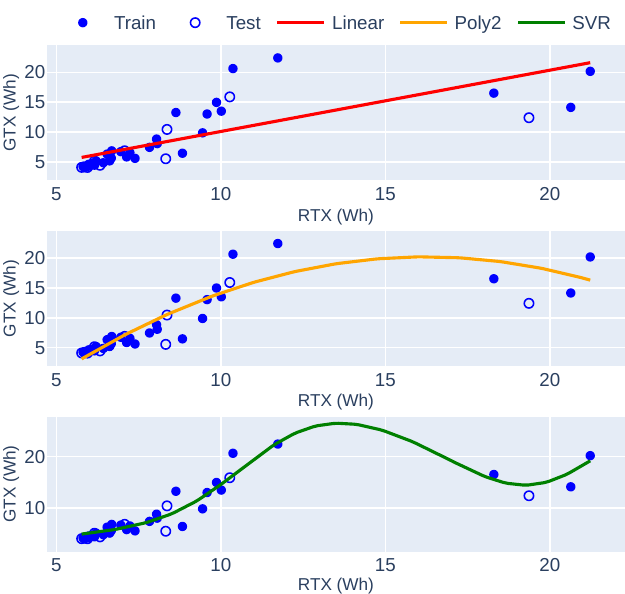}
    \caption{Illustration of different regression types to model the energy consumption of the RTX-GTX hardware pair.}
    \label{fig:RTX_GTX_regression_type}
\end{figure}

\begin{figure}[t]
    \centering
    \includegraphics[width=\linewidth]{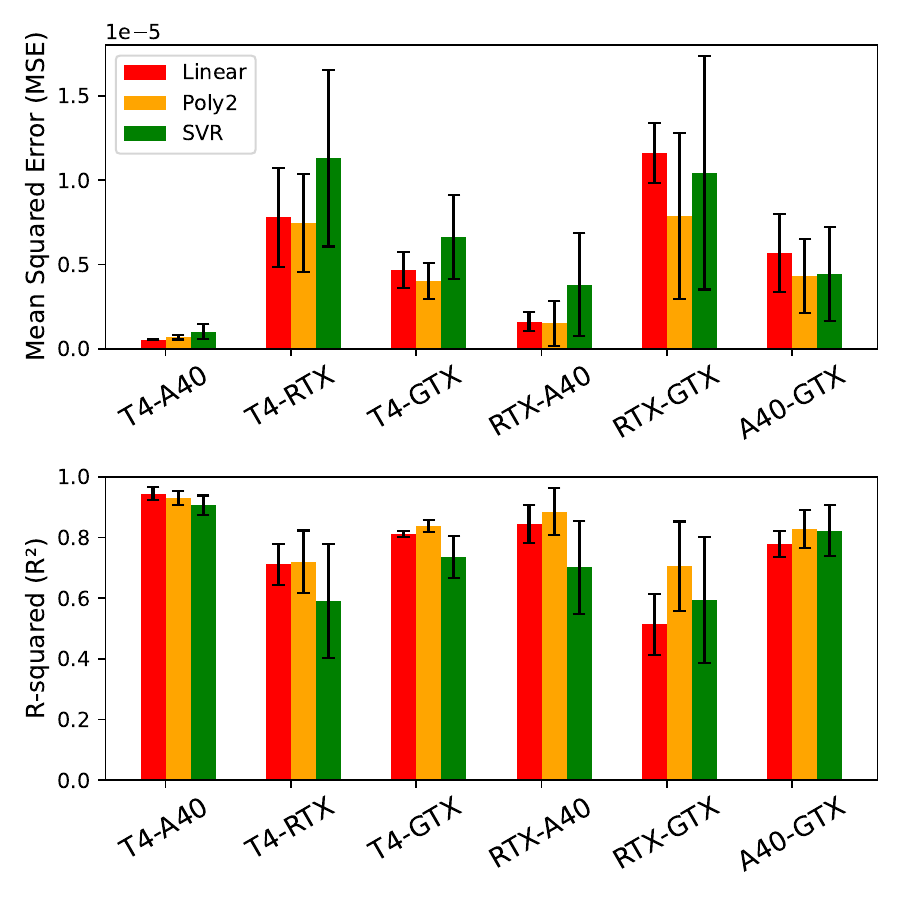}
    \caption{Comparison of regression types among linear, polynomial and support vector regression for normalizing energy consumption across different hardware.}
    \label{fig:regression_type}
\end{figure}

To quantify the impact of increasing the number of reference points, we divide our dataset of 43 models into two subsets: 80\% (34 models) for training and 20\% (9 models) for testing. This division is randomly generated five times to perform cross-validation. At each iteration, we select a range from 10\%, 15\%, 20\%, 50\%, and 100\% of the train data to be used as reference points, corresponding to 2, 4, 6, 16, and 32 models respectively. We explore two methods of subsampling reference models: a random selection of models and a min-max strategy, where we select half of the models with the lowest and half with the highest energy consumption, 10\% corresponding to our dual reference point approach. We assess the quality of the predictions by computing the coefficient of determination (R²) and the mean squared error (MSE). 

Results are presented in Figure \ref{fig:sampling}, where the left bar graph shows outcomes from random sampling (\ref{fig:random}) and the right from min-max sampling (\ref{fig:extrem}). We did not include the results of random sampling at 10\% due to its highly imprecise and poor normalization performance. Across all pairs, we see that increasing the percentage of the train set used as reference points for random sampling improves the R² and significantly reduces the MSE. For the T4-RTX pair, lower percentages such as 15\% and 20\% even result in negative R² values (not represented for more clarity) highlighting the ineffectiveness of random sampling at smaller sample sizes. Furthermore, while random sampling generally requires the entire train set to achieve accurate regression, a careful selection of just two points can be a viable strategy. Specifically, choosing two models at the extremes notably improves the results for the T4-RTX pair, outperforming the outcomes obtained using the full train set.

\paragraph*{Limitations.} The overall precision of the regression remains moderate for the T4-RTX pair. This suggests that the relationship between the energy consumption of the different devices may not be strictly linear and that further experiments are needed to explore alternative models that might better capture this relationship.

\subsection{Influence of the regression type}

In this section we explore different types of regression to model energy consumption across different hardware pairs. We compare linear, polynomial (degree 2) and support vector regression using 100\% of the training set.  For the SVR, we set the regularisation parameter $C=0.1$ and the loss tolerance $\epsilon=0.0001$ after a preliminary grid search.  An illustrative example of these regressions applied to a realisation of the train-test split is given in Figure \ref{fig:RTX_GTX_regression_type} for the RTX-GTX pair. We observe that increasing the complexity of the regression model tends to improve the fit to the training data points. Quantitative results of these observations are presented in Figure \ref{fig:regression_type}. For the T4-A40 pair, linear regression provides the best fit, achieving the highest R² and lowest MSE, indicating strong predictive accuracy. In contrast, polynomial regression outperforms linear regression for all other pairs, although with less confidence. SVR appears to be effective for the RTX-GTX pair but performs poorly for other pairings. 

\paragraph*{Limitations.} The performance of each regression model varies significantly depending on the hardware, making unified normalization across different types of hardware difficult. A preference for the linear regression emerges due to its computational efficiency and its generally robust performance and reliability in the majority of pairs. However, the variability observed in its performance across different hardware pairs suggests that focusing solely on energy might not fully capture the underlying dynamics of the energy consumption.

\begin{figure}[t]
    \centering
    \includegraphics[width=\linewidth]{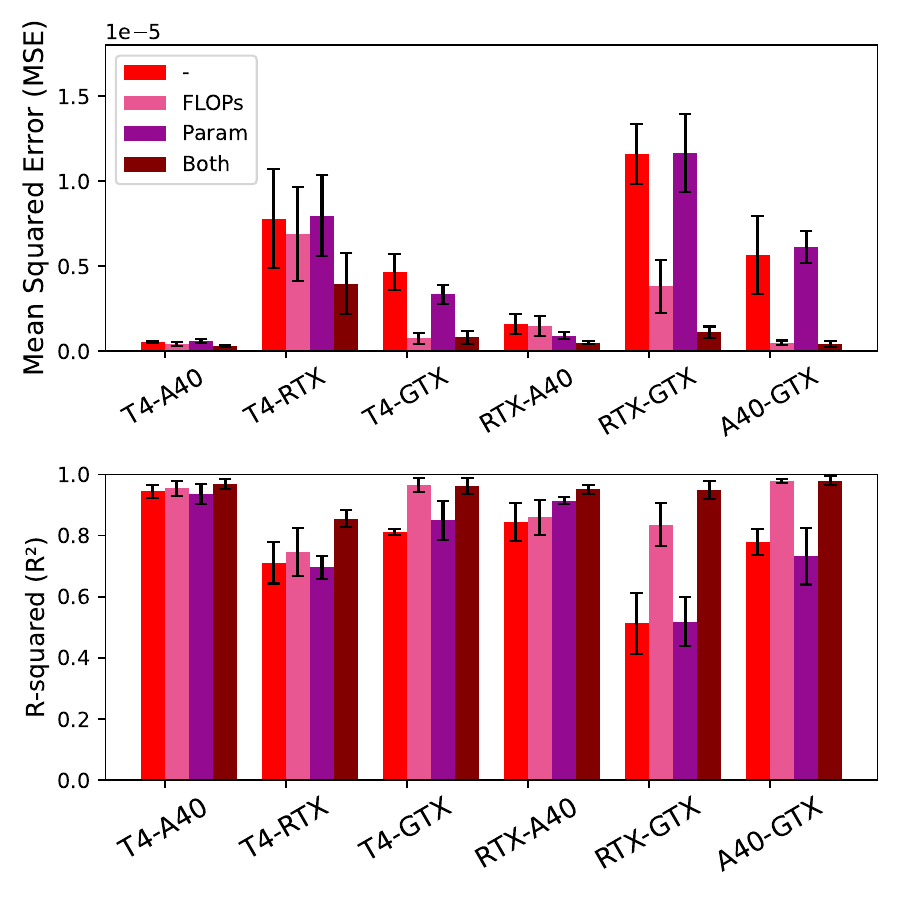}
    \caption{Comparison of linear regression models integrating additional computational metrics—FLOPs, number of parameters, and a combination of both—across various hardware pairs.}
    \label{fig:adding_conditions}
\end{figure}

\subsection{Influence of computational metrics}
\begin{figure}
    \centering
    \includegraphics[width=\linewidth]{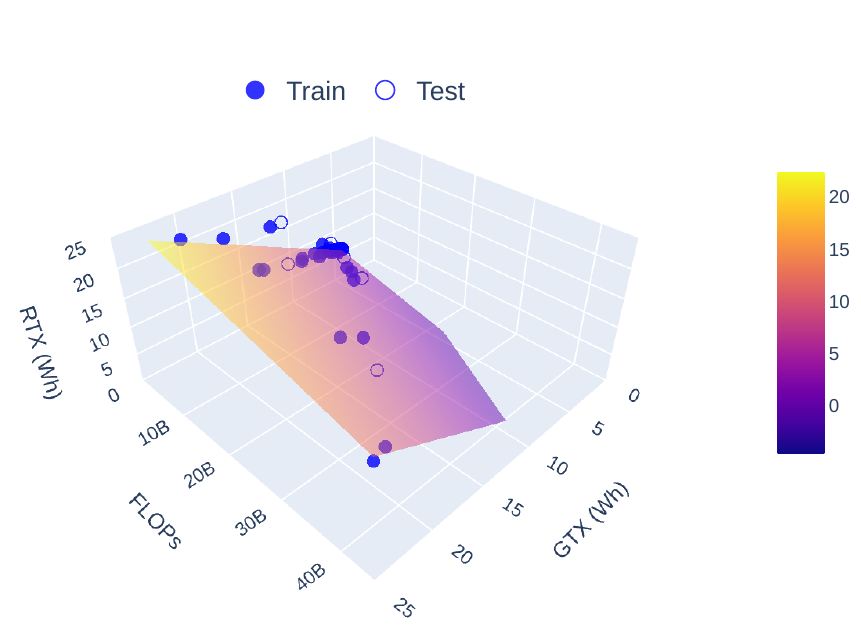}
    \caption{Illustration of a 3D linear regression model incorporating FLOPs alongside energy consumption data for the RTX-GTX hardware pair.}
    \label{fig:regression_3D}
\end{figure}

In this section we look at the effect of the incorporation of computational factors into linear regression models. Rather than examining the relationship between energy consumption and a 2D space, where each dimension represents the energy consumption of one hardware, we propose to extend the regression space by incorporating a third (and a forth) dimension, which is related to the architectural elements of the models. Therefore, we include the number of FLOPs, then the number of parameters, and finally both in the regression.
% \footnote{Additional figures of the 3D space can be found in the resource code.}
Figure \ref{fig:regression_3D} illustrates the linear regression when adding the FLOPs to the energy consumption for the RTX-GTX pair. We observe that the test data are closer to the regression surface compared to the 2D linear regression shown in Figure \ref{fig:RTX_GTX_regression_type}. This improvement suggests that incorporating FLOPs captures some of the nonlinear relationships from the 2D representations. 
Quantitative results of this analysis are presented in Figure \ref{fig:adding_conditions}. Across all hardware pairs, significant improvements in regression performance are noted when FLOPs are included along with energy. The inclusion of the number of parameters alone generally leads to decreased accuracy. Yet, combining both FLOPs and parameters yields even better results than using FLOPs alone for most hardware pairs. 
% Certain pairings exhibit specific trends: the A40-GTX pair demonstrate significant enhancements in predictive performance when FLOPs are included, either alone or combined with parameters. Meanwhile, the RTX-A40 pair exhibits the best performance progressively from FLOPs alone, to parameters, and finally to their combination.

\paragraph*{Limitations.} To normalize energy consumption using a three-dimensional approach with FLOPs, at least three reference points are required. When parameters are added, this requirement increases to four reference points, further complicating the normalization process and model selection.

\section{Conclusion and future works}

This paper presents several methods for normalizing the energy consumption of machine learning models across various types of hardware. We show that the choice of reference points has a critical impact on the normalization process. For example, the use of dual reference points provides a more balanced and accurate approach suitable for different hardware setups than using only one reference. Furthermore, we find that the integration of computational metrics such as the number of FLOPs and parameters improves the normalization strategy, but also introduces additional complexity.

Future work should address the complexity introduced by the inclusion of additional computational factors. In particular, efforts should focus on optimizing the balance between accuracy and practical applicability in real-world scenarios, ensuring that the normalization strategies developed are both effective and feasible for widespread use. In addition, it is crucial to extend the research to a wider range of machine learning applications, as well as different hardware configurations, neural network architectures and training routines.

\bibliographystyle{IEEEbib}
\section{References}
\ninept
\bibliography{refs}

\end{document}